\documentclass[11pt,a4paper]{article}
\usepackage[hyperref]{acl2021}
\usepackage{multirow}
\usepackage{graphicx} 
\usepackage{float}
\usepackage{times}
\usepackage{latexsym}
\usepackage{CJKutf8}
\usepackage{amsmath}

\usepackage{color}
\usepackage{color}

\makeatletter
\g@addto@macro\normalsize{%
  \abovedisplayskip 4pt plus 2pt minus 3pt%
  \belowdisplayskip \abovedisplayskip
  \abovedisplayshortskip 4pt plus2pt  minus3pt%
  \belowdisplayshortskip 4pt plus2pt minus3pt%
}

\usepackage{microtype}

\aclfinalcopy 

\title{Bridging the Gap Between Clean Data Training and Real-World Inference for Spoken Language Understanding}

\author{First Author \\
  Affiliation / Address line 1 \\
  Affiliation / Address line 2 \\
  Affiliation / Address line 3 \\
  \texttt{email@domain} \\\And
  Second Author \\
  Affiliation / Address line 1 \\
  Affiliation / Address line 2 \\
  Affiliation / Address line 3 \\
  \texttt{email@domain} \\}

\author{Di Wu$^\dagger$~~~~Yiren Chen$^\dagger$~~~~Liang Ding$^\ddagger$~~~~Dacheng Tao$^\ddagger$\\
$^\dagger$Peking University~~~~~~~~$^\ddagger$The University of Sydney\\
  {\tt inbath@163.com}~~~~{\tt yrchen92@pku.edu.cn}\\{\tt \{ldin3097,dacheng.tao\}@uni.sydney.edu.au}
  }
  
\date{}

\begin{document}
\maketitle
\begin{abstract}
Spoken language understanding (SLU) system usually consists of various pipeline components, where each component heavily relies on the results of its upstream ones.
For example, Intent detection (ID), and slot filling (SF) require its upstream automatic speech recognition (ASR) to transform the voice into text.
In this case, the upstream perturbations, e.g. ASR errors, environmental noise and careless user speaking, will propagate to the ID and SF models, thus deteriorating the system performance.
Therefore, the well-performing SF and ID models are expected to be noise resistant to some extent.
However, existing models are trained on clean data, which causes a \textit{gap between clean data training and real-world inference.}
To bridge the gap, we propose a method from the perspective of domain adaptation, by which both high- and low-quality samples are embedding into similar vector space. Meanwhile, we design a denoising generation model to reduce the impact of the low-quality samples. Experiments on widely-used dataset, i.e. Snips, and large scale in-house dataset (10 million training examples) demonstrate that this method not only outperforms the baseline models on real-world (noisy) corpus but also enhances the robustness, that is, it produces high-quality results under a noisy environment. The source code will be released.

\end{abstract}

\section{Introduction}

\begin{figure}[tp]
     \centering
     \includegraphics[width=0.48\textwidth]{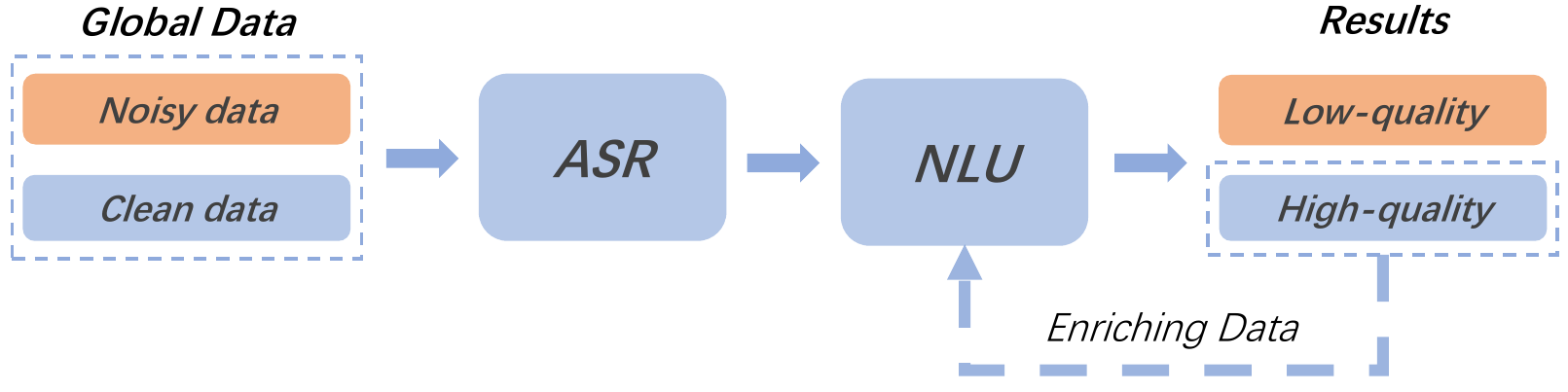}
     \caption{Illustration of pipeline components in SLU. Online high-quality samples are collected to enrich training corpus.
     }
     \label{fig-1} 
     \vspace{-8pt}
\end{figure}

In industry, personal assistants, such as Siri and Alexa, rely on a high-quality spoken language understanding (\citealp[SLU,][]{wang2005spoken}) system. Such a system usually contains many pipeline components, which serially depends on its upstream. As shown in Figure-\ref{fig-1}, an automatic speech recognition (\citealp[ASR,][]{yu2016automatic}) module transforms a utterance into text first and then natural language understanding (\citealp[NLU,][]{allen1988natural}) module extract corresponding slots and intents from that. After evaluation, the high-quality results will be utilized as enriching data to augment the NLU model. Due to its simplicity, this active learning pipeline is widely used in industry and academia.

However, such a system tends to be fragile since the errors will be gradually amplified when the information flow evolves in the pipeline. Complicated spoken environment, whether it's coming from the casualness of the speaker, the presence of environmental noise, or even the bias of the ASR system, can make the final analysis results be corrupted. Therefore, such noisy samples cannot be collected to enrich training data and data cleaning becomes an inevitable process for building an industrial SLU system. As depicted in Figure-\ref{fig-1}, only high-quality online results are collected as samples to further enrich training corpus. Although such active learning like method can eliminate noise from training corpus, it also causes another problem, that is, \textbf{a distribution gap between clean data training and online noisy inference}.

Recently, there are some researches focus on alleviating the impact of abnormal samples on the system from the perspective of robustness. \newcite{madry2017towards} study the adversarial robustness of neural networks through the lens of robust optimization. \newcite{dinan2019build} consider to defend adversarial human attacks in dialogue system by introducing an automatic approach to the “Build it Break it Fix it” strategy. \newcite{yasunaga2017robust} treat adversarial training as a regularization for the model, aiming to achieve robustness to input perturbations. \newcite{ding2020understanding} expose data-dependent priors derived from raw data (can be seen as noise data) into the training process on distilled data (can be seen as clean data) to bridge the distribution gap. \newcite{einolghozati2019improving} introduce and adopt adversarial training methods as well as data augmentation using back-translation~\cite{Sennrich:BT} to mitigate these issues. However, the abnormal samples or noise in these kind of methods are artificially designed and does not follow the distribution of the real-world noise.

In this paper, we propose to bridge the gap from the perspective of domain adaptation. In the training process, we incorporate both high-quality clean samples and low-quality noisy samples into training process. The noisy samples can be successfully embedded into a similar vector space with the clean data, and finally, be parsed by a denoising encoder-decoder slot parsing model (\S\ref{sec:method}). Experiments in \S\ref{sec:exp} demonstrate that our model not only promotes the performance globally but also mitigates the influence of abnormal samples, even if the error occurs inside the slots.
Our main contributions are:
\begin{itemize}
\item To the best of our knowledge, this is the first work to consider the distribution gap between clean data training and real-world online inference in SLU.
\item We propose a novel model for slot filling task, which has a significant advantage over the former SOTA models on noisy real-world data.
\item Our method can generate high-quality slots, even if the noise happen inside slot chunk. Traditional models do not have such capacity.
\end{itemize}

\section{Task and Data}
\label{sec:task-data}
\paragraph{Task}
As aforementioned, this paper aims to bridge the gap between \textit{clean data training} and \textit{real-world inference}. This problem is common in all kinds of industrial machine learning tasks.
Here we take one of the challenging SLU task -- slot filling (SF) for the specific case to show the superiority of our approach. Note that our approach is not limited to SF tasks, and we will explore it in future work.
In slot filling (SF), for an utterance like “Buy an air ticket from Beijing to Seattle”, SF task focus on words-level to figure out the departure and destination of that ticket are “Beijing” and “Seattle”. Accurate slot recognition is at the core of questions understanding.

\paragraph{Datasets}
Real-world data cleaning usually involves series of complex procedures, such as using language models and manually designed rule-based parser to detect and clean up noisy sentences. For example, as the winning submission in WMT2019\footnote{\url{http://www.statmt.org/wmt19/}}, \newcite{ding2019university} carefully design eight data cleaning features to construct the high quality corpus, making their models competitive in official test set. In real scenes, however, the user's input is often noisy. Therefore, merely considering improves the quality of the training set may be sub-optimal.

To investigate this gap in depth, we conduct experiments on both in-house real-world dataset and a widely-used dataset - Snips~\citep{coucke2018snips}. Our in-house dataset includes ~8M clean data collected from online system and ~2M noisy samples respectively, where the noisy part are detected and filtered by our in-house data cleaning pipeline. There are 65 slot labels and ~48K distinct words in this corpus. The detailed statistic can be found in Table-\ref{table-2}. For Snips dataset, to generate the artificially perturbed dataset, we employ common noise-injection practice to generate the noisy sentences~\cite{edunov2018understanding,ding2020self,liu2021noisy}. Specifically, we randomly sample N\% of the samples and processed them through three kinds of manually designed noise disturbance strategies\footnote{\url{https://github.com/valentinmace/noisy-text}}, that is, removing, replacing, or nearby swapping one time for a random word with a uniform distribution in a sentence. In particular, we empirically set the N as 20\% to keep the same noise ratio with our in-house dataset. Also, we avoid the manual disturbance happen in the slot chunk, which is the main difference with the in-house dataset.

\begin{figure*}[htp]
     \centering
     \includegraphics[width=1\textwidth]{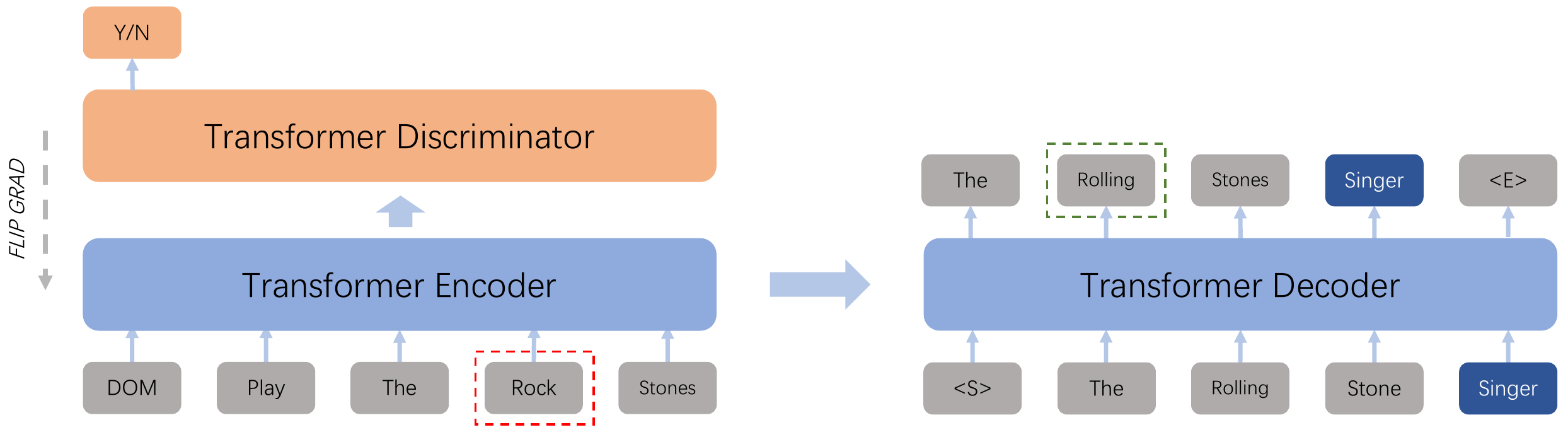}
     \caption{Illustration of RoSLU, where the left and right part indicate the domain adaptation and slot generation model respectively. A noise sample with in-slot errors in \textcolor[RGB]{217,117,137}{red} dotted box is mapped into a correct version to fool the discriminator. The slot value in \textcolor[RGB]{116,186,41}{green} dotted box are corrected , and final result is generated by the transformer based auto-regressive decoder. Note that slot label \textit{Singer} is treated as the same with the corresponding slot value.}
     \label{fig-2} 
\end{figure*}

\section{Proposed Method}
\label{sec:method}
In this section, we first describe how we bridge the gap between clean and noisy corpus from a perspective of domain adaptation. And then, we further introduce a denoising slot generation model, by which high-quality slot results can be parsed successfully even if the representation of a sentence are domain transferred. 

The brief schema of our model, named RoSLU, is shown in Figure-\ref{fig-2}, and details can be found in the corresponding caption.

\subsection{Adversarial Domain Adaptation}
\label{subsec:domain-adaptation}
To ensure the noisy samples can also be successfully parsed by the model trained on clean data, we introducing domain adaptation approach~\citep{ganin2016domain,zhang2019bridging}. As domain adaptation suggested, for bridging the gap between different domains, predictions should be made based on features that cannot discriminate between the training (clean) and test (noisy) domains. 

We follow the adversarial training strategy and introduced a transformer ~\cite{vaswani2017attention} based feature encoder $G_{enc}(\cdot;\theta_{enc})$ and domain discriminator $D(\cdot;\theta_{dis})$ respectively. Given a clean utterance $x\in S_{clean}$ and a perturbed utterance $\tilde{x} \in S_{noicy}$, we expect that, the encoded representations $G_{enc}(x;\theta_{enc})$ and $G_{enc}(\tilde{x};\theta_{enc})$ are in similar vector space so that can not be discriminated by $D(\cdot;\theta_{dis})$ correctly. To achieve this, we introduction an adversarial objective as follow: 
\begin{equation}
    \begin{split}
        &\mathcal{L}oss_{adv}(x, \theta_{dis}, \theta_{enc})\\
        & = \sum_{x\in S_{clean}}[-\log(D(G_{enc}(x, \theta_{enc}),\theta_{dis}))] \\
        & + \sum_{\tilde{x}\in S_{noisy}}[-\log(1-D(G_{enc}(\tilde{x}, \theta_{enc}),\theta_{dis}))]
    \end{split}
\end{equation}

An additional \textit{DOM} token is added at the beginning of the utterance for discriminator to predict whether the sample are noisy or clean. The discriminator aims to maximize $D(G_{enc}(x;\theta_{enc}), \theta_{dis})$ to 1 and minimize $D(G_{enc}(\tilde{x};\theta_{enc}), \theta_{dis})$ to 0, so that minimize the $Loss_{adv}$. While the feature encoder $G_{enc}(\cdot;\theta_{enc})$ are trained to maximize the loss function to fool the discriminator. The training procedure can be regarded as a min-max two-player game. Specificlly, we implement such procedure by flipping\footnote{multiplying by -1.} the gradients of the parameter $\theta_{enc}$. 

\subsection{Denoising Slot Generation}
\label{subsec:generation-model}
Although, by applying domain adaptation method introduced in \S\ref{subsec:domain-adaptation}, clean and noisy data can be mapped into a similar space, the impact of noise has not yet been mitigated. Besides, more importantly, the real-word noise could happen inside the slot chunks, so that the traditional sequential tagging model does not have the capacity to restore the corresponding slot correctly, as the perturbed utterance depicted in Figure-\ref{fig-2}.

Here, we introduce a transformer based auto-regressive generation architecture to handle this problem. Different from traditional sequential tagging model, we aims to generate slot name and slot label directly, where names and labels are treated as the same tokens. We convert original samples into the following parallel format: \textit{"play the rolling stones' love in vain ---- the rolling stones' SINGER love in vain SONG"}. 


\begin{table*}[htb]
\centering
\scalebox{1}{
\begin{tabular}{l|ccc|ccc}
\hline\hline
\multirow{2}{*}{Model} & \multicolumn{3}{c|}{\textbf{Snips Dataset}} & \multicolumn{3}{c}{\textbf{In-house Dataset}}

\cr\cline{2-7} & Clean & Noisy & Global & Clean & Noisy & Global \cr
\hline
Slot-Gated (w/o noisy)~\cite{goo2018slot} & 78.46 & 73.32 & 75.51 & 94.31 & 92.31 & 93.43  \\
SlotRefine (w/o noisy)~\cite{wu2020slotrefine} & 81.98 & 76.33 & 77.49 & 96.42 & 94.61 & 95.19 \\
\textbf{RoSLU (w/o noisy)} & 80.28 & \bf76.75 & 76.61 & 96.13 & \bf94.76 & 95.11 \\
\hline
Slot-Gated (w/ noisy)~\cite{goo2018slot} & 87.33 & 83.14 & 84.55 & 93.45 & 92.61 & 93.30  \\
SlotRefine (w/ noisy)~\cite{wu2020slotrefine} & 91.71 & 84.84 & 87.23 & 95.55 & 94.89 & 95.25 \\
\textbf{RoSLU (w/ noisy)} & 91.24 & \bf90.30 & \bf91.06 & \bf96.01 & \bf95.67 & \bf95.89 \\
\hline
\hline
\end{tabular}}
\caption{\label{table-1}Performance comparison between RoSLU and baselines on Snips and in-house datasets. Each experiment is conducted on both clean and noisy data to highlight the difference between clean training and real-world inference. \textit{Global} means merging clean and noisy testset, which reflects the real data distribution for in-house data.}
\end{table*}

Given an input utterance $\textbf{x} = x_1, x_2,...,x_M$, we directly optimize the generation probability of the corresponding slot sequence $\textbf{y} = y_1, y_2,...,y_n$:
\begin{equation}
p(\textbf{y}|\textbf{x};\theta_{enc},\theta_{dec}) = \prod_{n=1}^{N}P(y_n|\textbf{y}_{<n},x;\theta_{enc},\theta_{dec})
\end{equation}

where $\textbf{y}_{<n}$ is a partial generation results. $p(\textbf{y}|\textbf{x};\theta)$ is defined on a holistic transformer based neural network, which mainly includes two components: an  encoder shared with the feature encoder
$G_{enc}(\cdot;\theta_{enc})$ and another transformer based decoder with parameter $\theta_{dec}$ generates the \textit{n}-th target word based on the sequence of hidden representations. The standard training objective is to minimize the negative log-likelihood of the training corpus $S$ as follow:
\begin{equation}
    \begin{split}
        & \mathcal{L}oss_{s2s}(\textbf{x}, \textbf{y};\theta_{enc},\theta_{dec})\\ 
        & = -\sum_{x,y \in S_{clean}}\log P(\textbf{y}|\textbf{x};\theta_{enc},\theta_{dec})
    \end{split}
\end{equation}

\subsection{Training}
We update the parameters of both the adversarial domain adaptation model and slot generation model simultaneously as each iteration, rather than the periodical training strategy which is commonly used in adversarial learning.

Formally, given a training corpus $S_{global}$ merged from clean corpus $S_{clean}$ and noisy corpus $S_{noisy}$, the final training objective is:
\begin{equation}
    \begin{split}
       \mathcal{L}oss = \mathcal{L}oss_{s2s} + \alpha \cdot \mathcal{L}oss_{adv}
    \end{split}
\end{equation}
where $\alpha$ is a hyperparameter to balance adversarial loss and generation loss. We use SGD~\cite{zinkevich2010parallelized} to optimize our model. Notably, only clean samples can contribute generation loss, and it ensures the reliability of generative model.

\section{Experiments}
\label{sec:exp}
\paragraph{Baseline}
We chose two representative competitive models as our baselines:
\begin{enumerate}
    \item \textsc{Slot-Gated}\footnote{\url{https://github.com/MiuLab/SlotGated-SLU}}~\cite{goo2018slot}: they model the slot filling problem based on a bidirectional long short-term memory (BiLSTM) model~\cite{mesnil2015using}, where a gated mechanism is applied to capture the relationship from intent to slots.
    \item \textsc{SlotRefine}\footnote{\url{https://github.com/moore3930/SlotRefine}}~\cite{wu2020slotrefine}, where they joint model intent detection and slot filling task in a non-autoregressive fashion and introduce a two-pass iteration mechanism to capture the label dependence, achieving the SOTA performance with extremely low latency.
\end{enumerate}
   We conduct both clean and noisy environment evaluation based on their implementations. F1-score are used to measure the performance.

\paragraph{Setup}
All embeddings are initialized with xavier method~\cite{glorot2010understanding}. The batch size and learning rate are set to \{32, 0.001\} and \{1024, 0.01\} for Snips and the in-house datasets. Meanwhile, we set number of Transformer layers, attention heads and hidden sizes to \{4,8,96\} and \{4,8,256\} for these two corpus respectively. The hyper-parameters $\alpha$ is tested by grid search from 0.0 to 1.0 at 0.1 intervals.

\paragraph{Results of Noisy Inference}
As seen in Table-\ref{table-1}, upper part (``w/o noisy'') summarizes the model performance trained on cleaned in-house and snips corpus, while the lower part denotes the performance trained on global (containing noisy and cleaned) corpus.
As shown, the noisy inference performance on both Snips and in-house datasets can be significantly and consistently improved. In particular, our approach outperforms two strong baselines in both ``w/o noisy'' and ``w/ noisy'' settings. 

For clean training (``w/o noisy''), the better performance on the noisy test set of our RoSLU shows that it can resist the noise in user input.
For global training (``w/ noisy''), it is encouraging that our RoSLU outperforms two strong baselines over around 6 and 2 points in Snips and in-house test sets, respectively. This indicates that our method can effectively resist noise and improve the robustness of the model in real scenes.

\paragraph{Results on Clean Inference}
For clean inference, we report the model performance on both clean (``w/o noisy'') and global (``w/ noisy'') training settings. As shown in Table-\ref{table-1}, on Snips dataset, the clean inference results are 80.28 and 91.24, which are comparable to the best baselines (81.98 and 91.71); On in-house dataset, the clean inference results are 96.13 and 96.01, rivalling the best baselines (96.42 and 95.55).

To summarize, our RoSLU can achieve comparable performance with the best baseline model under clean inference setting, showing the generalization of our proposed approach.

\paragraph{Results on Global inference}
For the experiment on global data, both clean and noisy inference are considered. We also present the results from two perspectives: clean training and global (with noise) training.

For clean training setting (upper part in Table-\ref{table-1}), the global inference performance is similar to clean inference, where the RoSLU achieves comparable results to its counterpart baselines.
For global training setting (lower part in Table-\ref{table-1}), we find that the performance of RoSLU consistently outperforms baseline models on both Snips and in-house corpus. Particularly, on Snips dataset, RoSLU could outperform the best baseline model by +3.8 F-1 points, demonstrating its effectiveness and robustness. Even on our in-house data, RoSLU still achieves nearly +0.64 point of gain in the case of very sufficient samples and a high baseline score.

It is worth noting that after introducing the noisy corpus (``w/ noisy''), the performance on Snips are consistently improved, while the opposite trend exists on the inhouse dataset. The possible reason is that after cleaning up, the OOV problem becomes extremely serious due to the small volume of Snips and noise can function as a natural regularizer. However, for the in-house data with tens of millions of samples, this problem is negligible.

\begin{table}[tp]
\centering
\scalebox{1}{
\begin{tabular}{lrr}
\hline\hline
\textbf{Data} & \textbf{Clean} & \textbf{Noisy} \cr
\hline
Training data & 8M & 2M \\
validation data & 8K & 2K \\
Test data & 8K & 2K \\
\hline
Sampled noisy data & - & 500 \\
In-slot error & - & 177 \\
In-slot correction & - & 59 \\
\hline
\hline
\end{tabular}}
\caption{\label{table-2} Details of in-house data. We sample noisy test samples and manually evaluate the parsing results of them to verify the model capacity of slot correction. }
\end{table}

\paragraph{Analysis on In-Slot Error Correction}
For the in-house corpus, distubance may occur inside slot chunks, namely in-slot error. As shown in the Figure-\ref{fig-2}, the band name \textit{the rolling stones} is identified as \textit{the rock stones}, which may stemming from ASR error or careless speaking. 

Intuitively, the samples with in-slot errors and the similar clean samples can be mapped into the vector space, such that they have the potential to be successfully corrected by the generator. To verify this hypothesis, we sampled 500 samples from the in-horse test data and conduct the objective evaluation manually. As shown in Table-\ref{table-2}, we found that such in-slot disturbance can be found in 177 samples, and 59 samples of them are successfully corrected (both partial and overall correction in a utterance is counted), confirming our hypothesis. To further improve the in-slot error correction rate, we can design a specific module or procedure to explicitly alleviate it. We leave this as future work.

\section{Conclusion and Future Works}
In this paper, we reveal a real-world problem for SLU task, that is, there exists a distribution gap between training and inference. We show that this problem is not limited to SLU tasks, but is common in all industrial machine learning scenarios.

To address it, we present a novel adversarial training based slot generation model, named RoSLU, which significantly outperforms the traditional sequence tagging model in real-world scenario. 
Extensive experiments show that our proposed RoSLU could consistently outperform baseline models on noisy inference setting, keeping the comparable performance under the clean inference setting.
Further analysis shows that our model has the great potential to correct the in-slot disturbance, which usually happens in the real-world and can not be handled by traditional tagging models. 

For future works, we plan to extend our approach on other tasks, e.g., machine translation~\cite{vaswani2017attention} and cross-lingual word alignment~\cite{wu2021slua}. Also, it will be interesting to avoid in-slot errors by assigning more weights on the local tokens~\cite{luong2015effective,ding2020context}.

\bibliographystyle{acl_natbib}
\bibliography{acl2021}


\end{document}